\documentclass[journal,twoside,web]{ieeecolor}
\usepackage{array}
\usepackage{float}
\usepackage{longtable}
\usepackage{generic}
\usepackage{cite}
\usepackage{amsmath,amssymb,amsfonts}
\usepackage{algorithmic}
\usepackage{graphicx}
\usepackage{textcomp}
\usepackage{multirow}
\usepackage{url}

\def\BibTeX{{\rm B\kern-.05em{\sc i\kern-.025em b}\kern-.08em
    T\kern-.1667em\lower.7ex\hbox{E}\kern-.125emX}}
\begin{document}
\title{Transfer Learning in Electronic Health Records through Clinical Concept Embedding}
\author{Jose Roberto Ayala Solares, Yajie Zhu, Abdelaali Hassaine, Shishir Rao, Yikuan Li, Mohammad Mamouei, Dexter Canoy, Kazem Rahimi and Gholamreza Salimi-Khorshidi
\thanks{This research was funded by the Oxford Martin School (OMS) and supported by the National Institute for Health Research (NIHR) Oxford Biomedical Research Centre (BRC). It also got support from the PEAK Urban programme, funded by UK Research and Innovation's Global Challenge Research Fund, Grant Ref: ES/P011055/1. The views expressed are those of the authors and not necessarily those of the OMS, the UK National Health Service (NHS), the NIHR or the Department of Health and Social Care.}
\thanks{All authors are affiliated with Deep Medicine, University of Oxford, United Kingdom.}
\thanks{A.H., D.C. and K.R. are also affiliated with NIHR Oxford Biomedical Research Centre, Oxford University Hospitals NHS Foundation Trust, Oxford, United Kingdom.}
\thanks{A.H. (abdelaali.hassaine@wrh.ox.ac.uk) is the corresponding author.}}

\maketitle

\begin{abstract}
Deep learning models have shown tremendous potential in learning representations, which are able to capture some key properties of the data. This makes them great candidates for transfer learning: Exploiting commonalities between different learning tasks to transfer knowledge from one task to another. Electronic health records (EHR) research is one of the domains that has witnessed a growing number of deep learning techniques employed for learning clinically-meaningful representations of medical concepts (such as diseases and medications). Despite this growth, the approaches to benchmark and assess such learned representations (or, embeddings) is under-investigated; this can be a big issue when such embeddings are shared to facilitate transfer learning. In this study, we aim to (1) train some of the most prominent disease embedding techniques on a comprehensive EHR data from 3.1 million patients, (2) employ qualitative and quantitative evaluation techniques to assess these embeddings, and (3) provide pre-trained disease embeddings for transfer learning. This study can be the first comprehensive approach for clinical concept embedding evaluation and can be applied to any embedding techniques and for any EHR concept.
\end{abstract}


\section{Introduction}
\label{sec:introduction}
\IEEEPARstart{E}{lectronic} health records (EHR) are becoming the ubiquitous approach to keep track of an individual's health journey; they provide a comprehensive view into one's health record and include ``concepts'' such as demographic information, diagnoses, medications, laboratory tests and results, medical images, clinical notes, and more \cite{birkhead2015uses}. As a result, the availability of large EHR datasets has enabled a broad range of new applications in clinical informatics and epidemiology \cite{botsis2010secondary,jensen2012mining,shickel2017deep} including medical concept extraction \cite{meystre2008extracting,jiang2011study}, disease and patient clustering \cite{doshi2014comorbidity,li2015identification}, patient trajectory modelling \cite{ebadollahi2010predicting}, disease prediction \cite{zhao2011combining,austin2013using}, and data-driven clinical decision support \cite{kuperman2007medication,miotto2015case}, to name a few.

Recent developments in machine learning (ML), on the other hand, have provided the field of EHR research with an opportunity to train powerful models for predictive modelling and risk prediction. An important step in training ML models on EHR is the representation of patients as input data for these models; EHR being a sequence of heterogeneous (and often non-numeric) concepts that are recorded in irregular intervals, makes this step a challenge. In traditional epidemiological research, individuals have been often represented by a limited number of commonly collected variables or features \cite{ayala2019long}. Unfortunately, such approaches rely on prior knowledge about each feature's relevance to, and the interaction among, the features given the task of interest; such a prior knowledge can be difficult to source. Furthermore, the appropriate features are likely to vary from task to task, which turns the feature engineering to a hard-to-scale task in medicine, where there are many diseases and problems for which one might need to use ML. The early applications of ML for EHR have tried to alleviate this by extracting more features, and showed some improvements; for instance, Rahimian et al. \cite{rahimian2018predicting} showed that improving the feature extraction in a statistical ML framework can outperform a well-known hospital readmission model \cite{hippisley2013predicting} by nearly 10\% (in terms of AUC).

Despite such improvements, given the high number of potential features that one can extract from EHR (reaching hundreds, if not thousands or more), the high risk of missing the important features remains a key weakness of manual feature extraction. This is the problem that deep learning (DL) and artificial neural networks have solved in other domains, through their use of many linear and nonlinear transformations of inputs, across multiple layers, and resulting in more useful representations (hence the names ``representation learning'' \cite{bengio2013representation} and ``distributed representations'' \cite{hinton1986learning}). Such low-dimensional representations have been shown to improve the performance of models in domains such as computer vision \cite{szegedy2016inception}, natural language processing (NLP) \cite{hirschberg2015advances,graves2013speech} and clinical informatics \cite{tran2015learning,miotto2016deep}.

In this study, we aim to provide the field with a number of qualitative and quantitative benchmarking approaches that could be considered for evaluating such learned representation (also referred to as ``concept embeddings'') and hence improve the quality of transfer learning (through shared concept embeddings that are appropriately assessed and benchmarked). The remainder of this paper is organised as follows: Section~\ref{sec:related_work} provides an overview of the existing representation-learning approaches in EHR research, as well as related benchmarking studies; Section~\ref{sec:materials_methods} introduces the data and methodology used in this study; results are presented in Section~\ref{sec:results}; and Section~\ref{sec:discussion} concludes the paper with further discussions and key conclusions.

\section{Related work}
\label{sec:related_work}
One of the earliest works for learning both concept vectors and patient vectors from EHR was Tran et al.'s $\mathtt{eNRBM}$ (EHR-driven non-negative restricted Boltzmann machines) \cite{tran2015learning}, which was shown to learn clinically-meaningful representations of concepts (i.e., both diagnoses and medications), and patients. They evaluated these embeddings (i.e., the mapping of non-numeric concepts to vectors of real numbers) using both visual assessment (i.e., showing how diseases that are close in the International Classification of Diseases (ICD) hierarchy appear close to each other in the vector space) and predictive modelling (i.e., showing that the use of these vector representations improves the accuracy of suicide-prediction models). In a similar work, Miotto et al.'s $\mathtt{DeepPatient}$ \cite{miotto2016deep} used a three-layer stack of denoising autoencoders to learn distributed representation of patients; they evaluated the resulting patient embeddings by showing that they can lead to superior predictions in a range of clinical predictions tasks, when compared to other forms of patient representations such as feature extraction. In a more recent work, Nguyen et al. \cite{nguyen2016Deepr} used a Convolutional Neural Network architecture called $\mathtt{Deepr}$ (i.e., Deep record); one of the main differences between $\mathtt{Deepr}$ and previous works was its ability to simultaneously learn patient representations and predict unplanned readmissions, for which it outperformed a logistic regression using Bag of Words (BOW) patient vectors as input.

In addition to patient vectors, taking inspirations from NLP (given the similarities between EHR and language, as both being sequences of non-numeric concepts), Choi et al. introduced Med2Vec \cite{choi2016multi}, for learning the embeddings for both visits and medical codes. When compared to techniques such as Skip-gram \cite{mikolov2013efficient}, GloVe \cite{pennington2014glove}, and stacked autoencoders \cite{miotto2016deep}, for predicting future medical codes and clinical risk groups, Med2Vec showed a better performance. Later, Choi et al. \cite{choi2016retain,choi2016doctor,choi2017using} extended their work by using a Recurrent Neural Network for simultaneously learning both vector representations and clinical predictions; in another extension of their earlier works, Choi et al. introduced GRAM \cite{choi2017gram}, a graph-based attention model that learns disease embeddings and combines them with hierarchical information inherent to medical ontologies. Both these improvements led to better performance in clinical prediction tasks.

Most of the works discussed so far do not fully take into account the temporal nature of EHR. In a recent work by Cai et al. \cite{cai2018medical}, authors proposed a time-aware attention model to address this issue; their model simultaneously learns representations and temporal scopes of medical concepts. They employed clustering and nearest neighbour search tasks to evaluate the quality of their medical concept embeddings and observed improvements over Continuous BOW (CBOW), Skip-gram, Glove and Med2Vec. In another related work, Xiang et al. \cite{xiang2019time} attempted to take time into account by extending three popular embedding techniques (Word2Vec \cite{mikolov2013distributed}, positive pointwise mutual information \cite{arora2016latent,beam2019clinical}, and FastText \cite{bojanowski2017enriching}) to consider time-sensitive information. They used clustering- and classification-based evaluation frameworks to show the improvements that resulted from their approach.

More recently, Transformer -- a new deep-learning architecture which does not rely on sequential processing of data and instead employs an attention mechanism to learn the interdependencies among various concepts in a sequence -- has been growing in popularity in representation learning. Transformers were shown to outperform most common alternatives in a broad range of tasks\cite{vaswani2017attention}. One of the most successful use cases of Transformer models was in ``BERT'' (Bidirectional Encoder Representations from Transformers), which has achieved state-of-the-art performance in many NLP tasks \cite{devlin2018bert}. The technique has also been applied to learn the latent EHR patterns; Graph Convolutional Transformer \cite{choi2019graph} and Med-BERT \cite{rasmy2020med} are two examples of such applications. One of the best performances of risk prediction in EHR was shown to result from a Transformer-based approach named BEHRT \cite{li2020behrt}; it outperformed previous deep learning architectures in predicting the on onset of many diseases. The success of such techniques is due to their ability to learn contextualised embeddings, and hence allowing a better representation of different clinical concepts and the overall sequence.

Overall, the use of representation learning and concept embeddings that result from them are fairly new and growing in medicine; new techniques are being developed to use the complexities found in healthcare data such as EHR as an advantage towards more accurate predictions. Nevertheless, the exploration of ways to compare, benchmark and assess the quality of different embeddings in EHR remains an under-investigated topic, despite the important role this plays in EHR DL research. Even in fields such as NLP, with longer history of transfer learning through multi-purpose word embeddings, there are works as recent as Chen et al. \cite{chen2018visual} that propose tools to explore and compare word embeddings from different training algorithms and textual resources, in order to identify clusters of words in terms of word embeddings, the semantic direction, and the relationship between semantically related words. In another example, Wang et al. \cite{wang2018comparison} perform a qualitative and quantitative analysis where they focus on a comparison of word embeddings trained from a Skip-gram model using different corpora (namely clinical notes, biomedical publications, Wikipedia, and news) to address biomedical natural language processing applications. There are also some benchmarking efforts, which aim to compare different methods in terms of natural language inference, recognising question entailment and question answering \cite{abacha2019overview}. Similar benchmarking exists for key phrases annotation in medical documents \cite{lara2019key}, for the evaluation of embeddings derived from clinical notes \cite{beam2019clinical} or for assessing how representative are embeddings for medical terminology \cite{schulz2020can}. All these methods evaluate embeddings resulting from textual information rather than structured EHR. To the best of our knowledge, no benchmarking approaches have been proposed for semantic evaluation of embeddings resulting from EHR data.

In this paper, we propose a set of approaches for both qualitative and quantitative evaluation of EHR concepts' embeddings, which will help improve the quality of transfer learning (through shared concept embeddings that are appropriately assessed and benchmarked) in the field. We consider five representation learning frameworks: Autoencoders, Neural Collaborative Filtering (NCF), Continuous Bag-of-Words (CBOW), CBOW with Time-Aware Attention (CBOWA), and BEHRT; each one of these approaches will be trained and validated for disease embedding on one of the world's largest and most comprehensive datasets of linked primary care EHR, known as Clinical Practice Research Datalink (CPRD). We believe the list of techniques we chose is representative of the most commonly used methods for representation learning in EHR, which have been shown to outperform some of their counterparts in mapping diseases to ``useful'' vectors (see Shickel et al. \cite{shickel2017deep}). In addition to the evaluation of the disease embeddings resulting from each of these methods, we share these embeddings with the field to encourage further research in this direction, as well as the use of the best ones -- according to our advocated benchmarking process -- for some downstream tasks, and promoting transfer learning in EHR research through concept embeddings. 

\section{Materials and Methods}
\label{sec:materials_methods}
\subsection{EHR Data}
The source of EHR data for this study was the UK Clinical Practice Research Datalink (CPRD) \cite{herrett2015data}, a service that collects de-identified longitudinal primary care data since 1985 from a network of GPs in the UK, which are linked to secondary care and other health and area-based administrative databases \cite{cprd}. These linked databases include the Hospital Episode Statistics, or HES (for data on hospitalisations, outpatient visits, accident and emergency attendances, and diagnostic imaging), the Office of National Statistics (death registration), Public Health England (cancer registration), and the Index of Multiple Deprivation. Patients included in the CPRD database are nationally representative in terms of age, sex and ethnicity. Given the data on demographics, diagnoses, therapies, and tests together with its linkage to other health-related databases, the CPRD is a valuable source of healthcare data \cite{herrett2015data}. Because CPRD contains detailed personal information, the dataset is not readily available to the public, and its usage depends on approval from the CPRD Research Ethics Committee \cite{cprd}.
In this study, we only considered practices providing healthcare data that met research quality standards within the period from 1 January 1985 to 31 December 2014, and agreeing for their patients' records to be linked to the Hospital Episode Statistics national database. Furthermore, we focused on patients aged 16 years or older who have been registered with their GP for at least 1 year, and that have at least 5 visits in their records. This resulted in a dataset of 3,092,631 patients, and is profiled in Table \ref{tab:stats}.

\begin{table}
\caption{A summary of CPRD dataset from 1 January 1985 to 31 December 2014 for the selected cohort of 3.1 million individuals. (SD: Standard Deviation, IQR: Interquartile Range)}
\label{table}
\setlength{\tabcolsep}{3pt}
\begin{center}
\begin{tabular}{|l|l|}
\hline
Number of patients& 
3,092,631\\
\hline
Number of visits (in GP or Hospital)& 
57,918,684\\
\hline
Number of visits per patient, Mean (SD)& 
18.73 (17.16)\\
\hline
Number of visits per patient, Median (IQR)& 
13 (17)\\
\hline
Number of disease codes& 
1899\\
\hline
Number of codes in a visit, Mean (SD)& 
1.36 (1.07)\\
\hline
Number of codes in a visit, Median (IQR)& 
1 (0)\\
\hline
\end{tabular}
\end{center}
\label{tab:stats}
\end{table}

While CPRD contains many data fields, in this study, we limited our analyses to diagnoses, in order to explore ``comorbid'' conditions. Some demographics variables were also taken into account including sex (binary), region (categorical with 10 classes), and birth year (categorical with 111 classes ranging from 1888 to 1998). In CPRD, diagnoses are coded in Read codes for primary care \cite{read_codes} and ICD-10 codes for HES \cite{icd10}. In their raw format, these coding schemes were unsuitable to work with given their high cardinality, i.e., there were around 110,000 Read codes and 14,000 ICD-10 codes. Many of the Read codes do not correspond to actual diagnoses (but rather to other information such as procedures, family medical history, occupations, and so on). In order to have a unified coding system, we mapped diagnoses from Read codes to ICD-10 codes using the mapping provided by NHS Digital \cite{read_icd}. Furthermore, we limited the ICD-10 codes to three-characters, resulting in approximately 1800 disease codes. Beyond the $3^{rd}$ character, other granular details such as the anatomic site or severity of the disease are provided but these are not always present and are mostly used for billing purposes. Moreover, further granularity can lead to lower frequency for many codes, which can decrease the quality of the learned representation for less frequent diseases; this might be the reason many such representation learning works in the field have operated at this level of granularity \cite{tran2015learning,nguyen2016Deepr}

\subsection{Representation Learning Methods}
\begin{figure*}[!t]
\centerline{\includegraphics[width=1\textwidth]{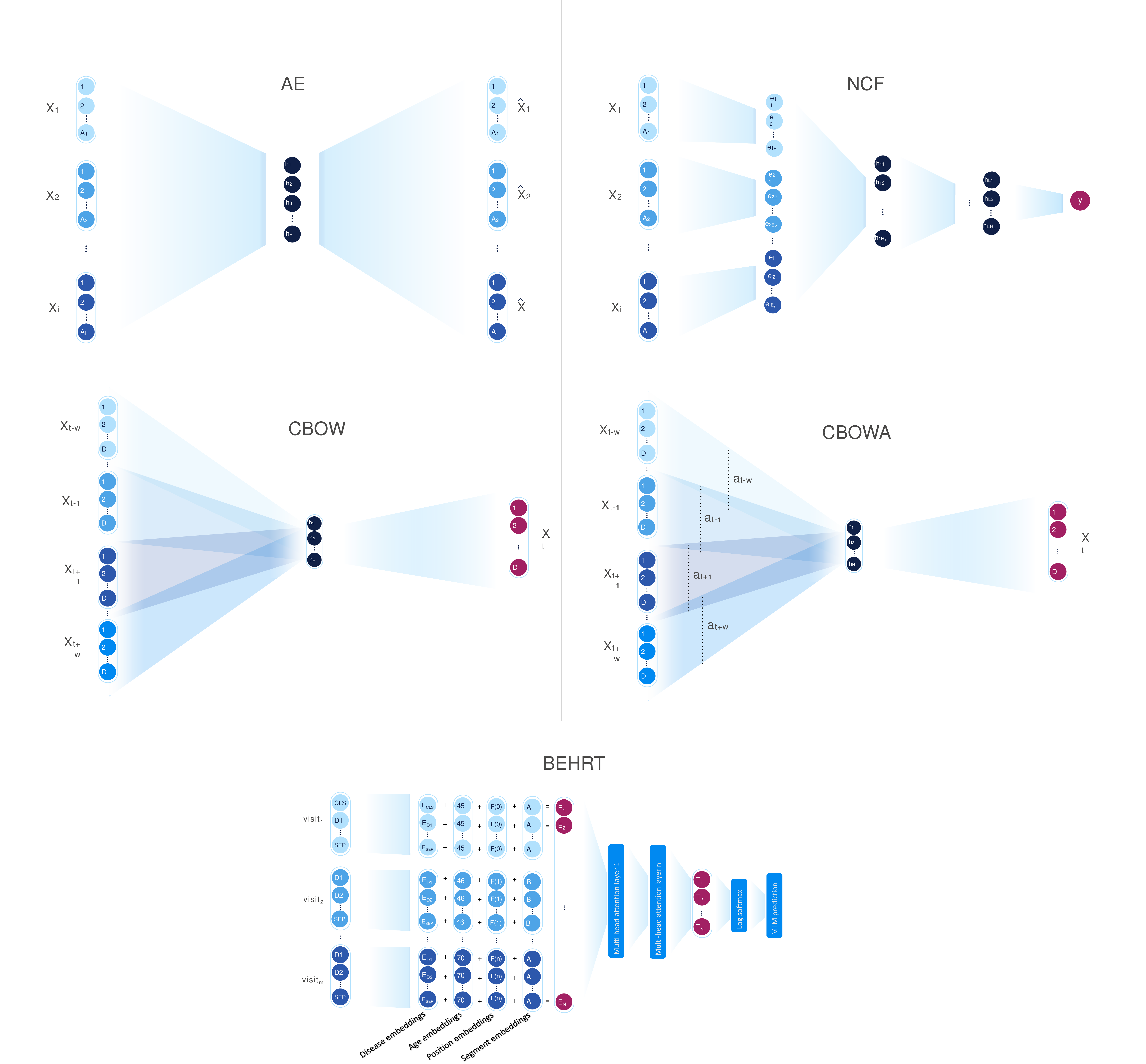}}
\caption{The five representation-learning architectures used in this paper. For AE and NCF (top row), the input consists of diseases and additional concepts such as sex and region; for CBOW and CBOWA (second row), we only used the diseases; BEHRT (bottom row) uses in addition to diseases, temporal information such as age at different encounters as well as position of different encounters.}
\label{fig:architectures}
\end{figure*}

In this section, we briefly introduce the five representation-learning methods that we employed for learning the disease embeddings. Our simplest architecture is an autoencoder (AE), a type of artificial neural network that can learn a new (and more efficient) representation of its inputs, in an unsupervised manner. As shown in Fig. \ref{fig:architectures}, AE is trained to learn a lower-dimensional representation of its input that can reconstruct the original input data as closely as possible. For this work, we based our AE model on the model architecture from Miotto et al. \cite{miotto2016deep} and the data format from Tran et al. \cite{tran2015learning}. Our model consisted of a single layer with 10 hidden units, a learning rate of 0.1 and a noise rate of 0.05 to train the AE for 7 epochs. Each patient's medical history was aggregated into a sparse vector, where each entry corresponded to the number of times a single disease was diagnosed. All categorical features (sex, region, birth year) were converted to dummy variables. This resulted in a 2,022 dimensional input.

Neural collaborative filtering (NCF) is the second framework that we used (see Fig. \ref{fig:architectures}). Based on the earlier work of He et al. \cite{he2017neural} and Howard et al. \cite{howard2018fastai}, our model architecture consisted of 3 hidden layers (with 100, 50 and 10 hidden units, respectively) and an output layer with a single classifier unit. In this case, our data consist of sex, region, birth year, age and disease codes, which were embedded into 1, 6, 22, 23, and 110 dimensions, respectively (these dimensions were computed using the rule-of-thumb formula from \cite{howard2018fastai}). In order to define a classification problem, we labelled the actual samples from the data as ``positive''; additionally, we generated ``negative'' samples by creating records that were not available in CPRD \cite{goldberg2014word2vec}. That is, for every positive record, we generated two negative records by randomly sampling from the age years and disease codes, and verifying that such new records did not exist in CPRD. We trained the NCF in this two-class classification task.

In addition to AE and NCF, we employed the continuous bag-of-words (CBOW) model, originally proposed by Mikolov et al. \cite{mikolov2013efficient}, which learns word/concept embeddings by using the context words within a sliding window to predict a target word (see Fig. \ref{fig:architectures}). This has been one of the standard approaches to learn embeddings in NLP. We used the implementation provided by Řehůřek et al. \cite{rehurek2010software} with the suggested hyperparameters to create embeddings of size 110. We trained the CBOW model by formatting any given patient's data as a sequence of disease codes, where disease codes are handled in the same way words are handled in NLP \cite{hirschberg2015advances,li2020behrt}.

Given the importance of temporal information in EHR, Cai et al. \cite{cai2018medical} modified the CBOW model to take such temporal characteristics into account, using an attention mechanism \cite{xu2015show,luong2015effective,britz2017massive,vaswani2017attention} that learns a time-aware context window for each disease code (see Fig. \ref{fig:architectures}). We trained the CBOWA model using the same code available at the authors' GitHub repository \cite{Repository_CBOWA}. The model creates embeddings of 100 dimensions using a learning rate of 0.01 and a negative sample of 5 for 10 epochs. We trained the CBOWA by formatting any given patient's data as a sequence of time-stamped disease codes, and treating disease codes in the same way words are treated in NLP.

Our final model is BEHRT, which is based on the Transformer architecture \cite{li2020behrt}. By depicting diagnoses as words, visits as sentences, and patients’ medical histories as documents, this architecture supports temporal information through the use of positional encoding as well as age and segment encoding. It was trained to predict masked disease tokens using 10 attention heads and 4 hidden layers (see Fig. \ref{fig:architectures}). For more details about the BEHRT model, please refer to the original paper \cite{li2020behrt}.

\subsection{Benchmarks and Evaluation}
\label{sec:benchmarks}
As discussed earlier, the quality of EHR research through medical concept embeddings (including transfer learning) will depend on the quality of such embeddings. Therefore, in this study, we aim to introduce a series of benchmarking approaches that can assess such embeddings and evaluate them for use for transfer learning. The simplest of our advocated approaches are qualitative assessments. In order to do so, we first created for each disease a list of closest diseases (using cosine similarity as the measure of distance); these neighbourhoods were then assessed by human experts. Next, we mapped these embeddings to a 2D space using t-SNE \cite{maaten2008visualizing}; the goal here is to compare the resulting disease clusters (based on embedding vectors' distances) with what we know from ICD-10 disease hierarchy.

Both these approaches are difficult to use at scale, and can suffer from variability, due to differing opinions of experts. Therefore, we next introduced a number of quantitative assessments, which score the similarities between the relationships among diseases in the embedding space and their relationships in some a priori known medical space. We used three different sources of medical knowledge as compiled in \cite{hassaine2019learning}, each providing disease pairs that are associated with each other, i.e., pairs of comorbid conditions: (1) Jensen et al. \cite{jensen2014temporal} concluded a list of 4,014 comorbid disease pairs based on the analysis of a large national health dataset, followed by thorough medical due diligence; (2) Dalianis et al. \cite{dalianis2009stockholm,comorbidity_view} studied the Stockholm EPR Corpus to conclude a list of 1,000 disease pairs; and (3) and Beam et al. \cite{beam2019clinical} studied a large body of medical literature to conclude a list of 359 comorbid condition pairs. Furthermore, through the study of large body of medical literature, they also concluded a list of 724 causal pairs, where one disease causes the other. When evaluating each embedding approach, we calculate the percent of disease pairs ($d_i$,$d_j$), where $d_i$ is in $d_j$'s L-neighbourhood (i.e., the closest $L$ diseases to a disease of interest, according to cosine similarity), or vice versa.

As the final benchmarking of our embeddings, we tested them in a number of downstream prediction tasks. The classification task we chose was whether a patient will develop a given disease within 6 months leading to his/her last recorded visit. The inputs to our model (i.e., a feed-forward neural network) were sex, region, birth year, and disease history for each patient. The first three input variables were considered as categorical features, while the disease history was formatted as a sparse vector, where each entry corresponded to the number of times a single disease was diagnosed for a patient. The architecture of our neural network consisted of three layers (with 100, 50, and 10 hidden units, respectively), with a single classifier output unit. To benefit from the transfer learning, we used the learned embeddings as the weights that connect the input layer to the first hidden layer. Note that, unlike AE and NCF, CBOW, CBOWA and BEHRT did not provide embeddings for sex, region, and birth year. Therefore, we made two separate evaluations using the embeddings obtained for sex, region, and birth year from the AE and NCF models with the disease embeddings from the CBOW, CBOWA and BEHRT models (see Table \ref{tab:prediction_results}). Furthermore, in order to assess the benefit of using these learned representations, we considered the case where the disease embeddings needed to be learned for each task, i.e., the end-to-end approach, instead of using transfer learning (i.e., the one we refer to as ``Random'').

Lastly, in addition to qualitative, quantitative, and prediction-based assessment, and in order to assess the robustness of the embeddings to run-to-run variability, and their sensitivity to the size of the training dataset, we carried out two reliability analyses. These will provide an additional assessment of the embeddings' quality in terms of reflecting the true meaning in the data, with lower risk of under-fitting and higher generalisation ability (see Antoniak et al. \cite{antoniak2018evaluating} for similar assessment in the NLP space). First, we checked the run-to-run variability by training the models 10 times and computing the confidence interval of the cosine similarity metrics for some disease pairs. In a second experiment, we checked the effect of sample size on the cosine similarities of all embeddings pairs, by training the models 10 times on 20\%, 40\%, 60\%, 80\% and 100\% of the available data, and computing the corresponding average standard deviation among all cosine similarities.

\subsection{Code and Embeddings availability}
For each method, we used the authors' original code; otherwise, we implemented them based on the descriptions in the papers. All our code is implemented in PyTorch 1.0.1 \cite{NEURIPS2019_9015}, and scikit-learn 0.20.1 \cite{scikit-learn}, and run on two NVIDIA Titan Xp graphics cards. Furthermore, all hyperparameters were manually tuned using the learning from our own previous related works \cite{solares2020deep} and the documentation that accompanied the authors' original code. In addition, we provide a downloadable set of pre-trained disease embeddings for other researchers to use, which will be available at https://github.com/deepmedicine/medical-concept-embeddings.

\section{Results}
\label{sec:results}
In this section, we show the results from various benchmarking approaches, as well as the reliability of the embeddings resulting from each representation-learning technique.

\subsection{Qualitative and Quantitative Assessment}
\begin{figure*}[!t]
\centerline{\includegraphics[width=0.9\textwidth]{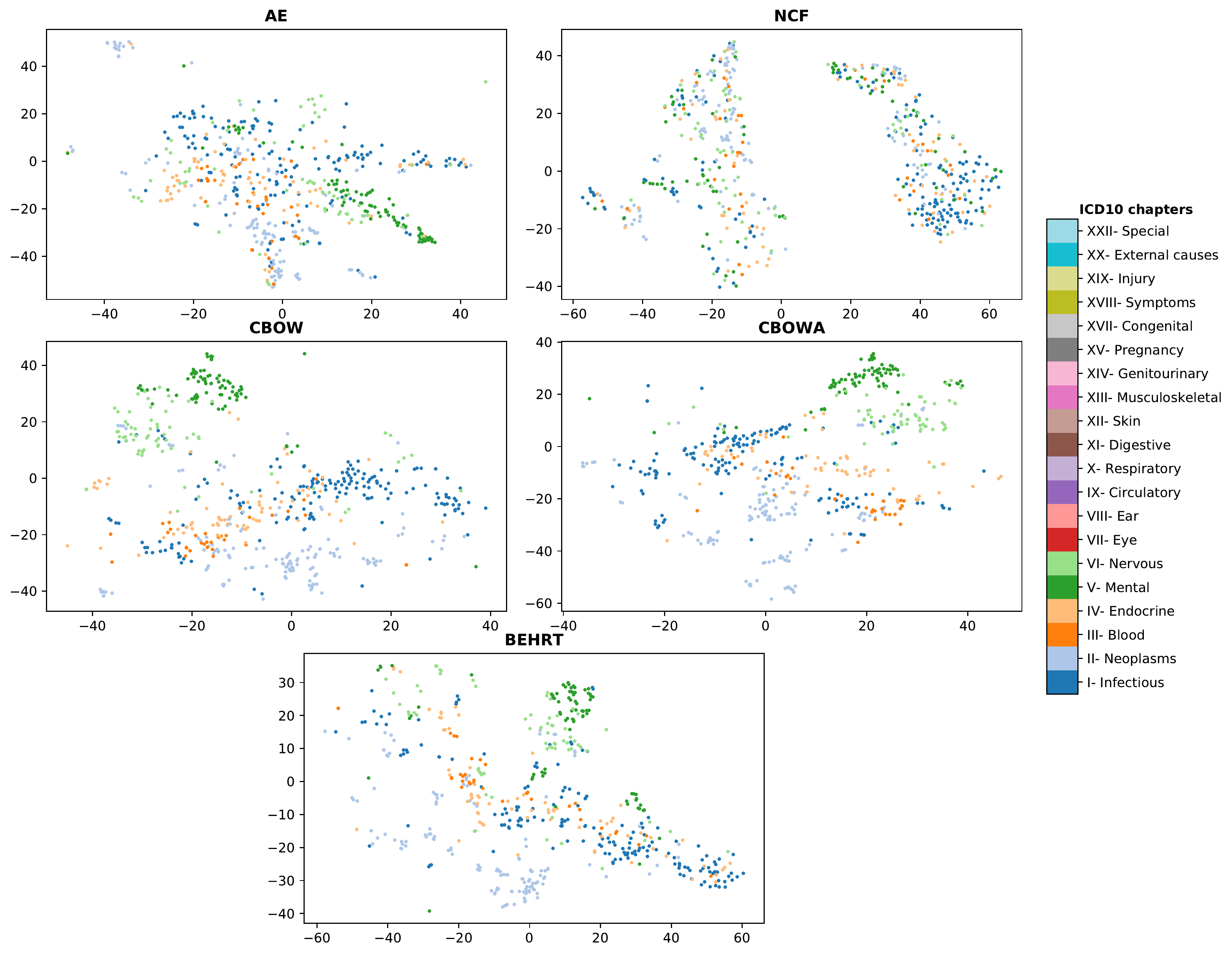}}
\caption{2D projection of the disease embeddings learned from AE, NCF, CBOW and CBOWA. Each dot denotes a disease and colours uniquely represent the ICD-10 chapters. Overall, the embeddings obtained with the CBOWA model seem to form more concise clusters, aligned with the colouring from a priori medical knowledge we have from the hierarchical structure of ICD-10. An online interactive version of this figure is available at \protect\url{https://deepmedicine.github.io/embeddings/}}
\label{fig:t_sne}
\end{figure*}

The t-SNE graphs in Fig. \ref{fig:t_sne} show the closeness of diseases in the embedding spaces; we assigned a unique colour to each ICD chapter to help the visual investigation. Embedding approaches where nearby diseases show similar colours can be seen as being in correspondence with {\em a priori} medical knowledge coming from ICD-10 disease hierarchy. Overall, the embeddings obtained from CBOWA and BEHRT seem to form more concise clusters that agree with the hierarchical structure of ICD-10. Furthermore, we chose some of the diseases and derived their 10 nearest neighbours (based on cosine similarity) for expert evaluation. The resulting lists are shown in the Tables in the Supplementary Information section. According to expert's evaluation, all the embeddings seem to capture meaningful associations often from different perspectives, although some patterns seem to be related to how diseases are recorded in the health system. BEHRT's inability to learn the appropriate representations for rare conditions (as can be seen in the most similar diseases to hypertension), is in line with what was previously shown \cite{li2020behrt}.

\begin{figure}[ht]
\centerline{\includegraphics[width=\columnwidth]{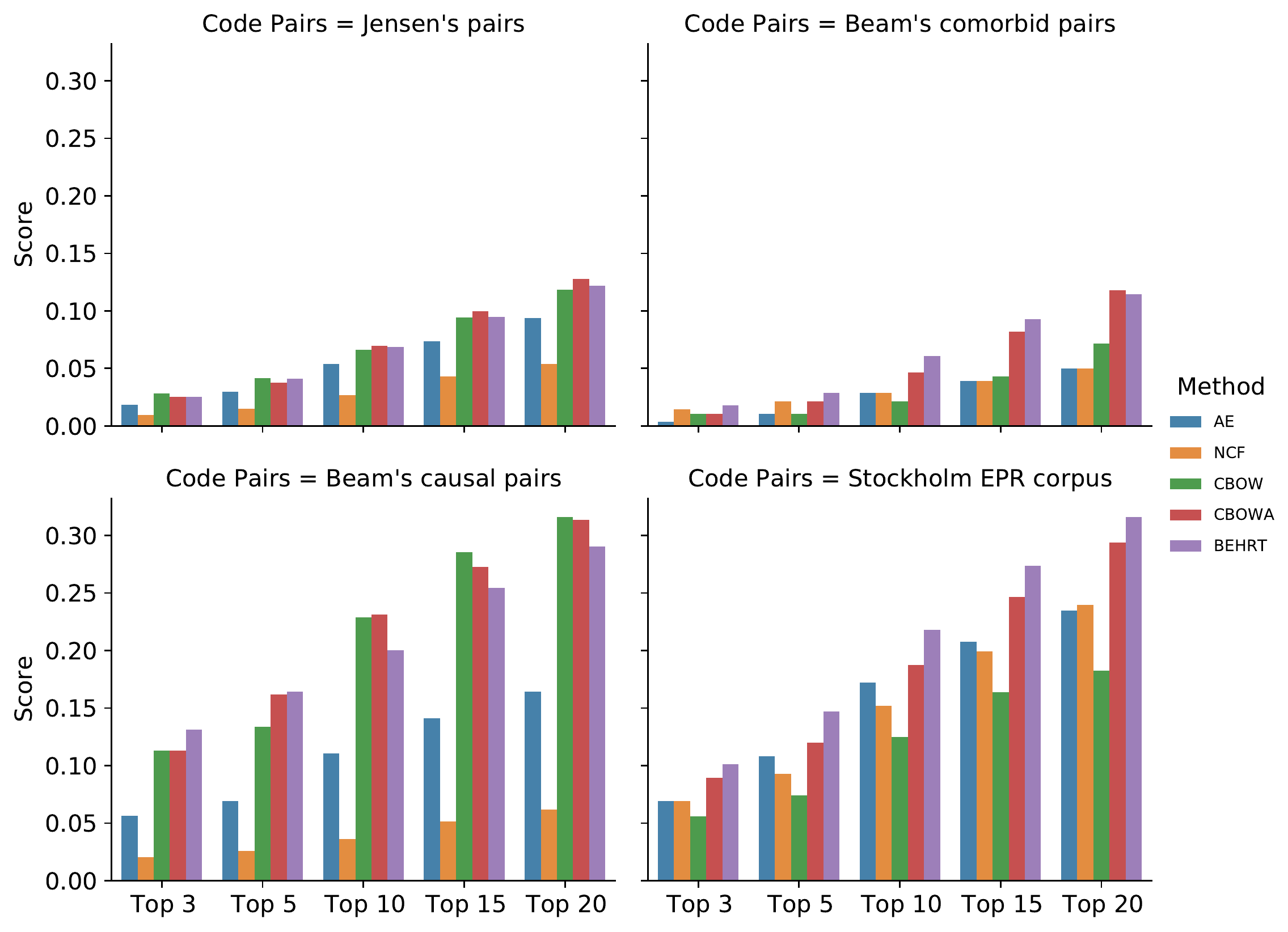}}
\caption{Neighbour analysis result based on the cosine similarity between the embeddings of different ICD codes.}
\label{fig:neighborhood}
\end{figure}

Following the qualitative assessment, we carried out quantitative benchmarking for neighbourhood sizes ranging from 3 to 20 diseases. For each neighbourhood, we calculated the percentage of diagnosis pairs that appeared within the neighbourhood of either of the diseases in that pair (see \ref{sec:benchmarks}). The results are shown in Fig. \ref{fig:neighborhood}, where we see that context-aware representation learning models are better at capturing statistically significant relationships among pairs of diseases. In Fig. \ref{fig:neighborhood}, we show the percentage of disease pairs ($D_1$,$D_2$) where $D_1$ (or $D_2$) is one of the k-nearest neighbours of $D_2$ (or $D_1$) using the cosine similarity metric. The value of $k$ varies from 3 to 20. These percentages are computed from different sources of disease pairs including the works of Jensen et al. \cite{jensen2014temporal}, Dalianis et al. \cite{dalianis2009stockholm,comorbidity_view} and Beam et al. \cite{beam2019clinical}.

\subsection{Prediction Task}
We carried out the prediction task for three different diseases; see Table \ref{tab:prediction_results} for more details. Despite us selecting the diseases with high prevalence (i.e., high percent of people in our data that had them), in all three classification tasks, the target variable was imbalanced (i.e., we had much lower number of positives than negatives). Therefore, we evaluated the models using average precision score (APS) and F1 \cite{sk_f1_score} metrics. Note that APS summarises a precision-recall curve as the weighted mean of precisions achieved at each threshold, with the increase in recall from the previous threshold used as the weight \cite{sk_avg_precision,Zhu2004Recall}. The prediction is performed using FastAI tabular learner where sex, region and year of birth are fed as categorical variables and every disease is fed as a continuous variable. Note that the aim of this downstream prediction task is to evaluate the performance of the tabular learner when the only information fed to it relates to the embeddings being evaluated. While some models have the ability to incorporate other information, such as the timing of diagnoses, in order to make the comparison as fair as possible, we only focused on disease, sex, region and year of birth inputs (the accuracy of performance due to the embeddings). Results show that BEHRT has the best performance overall. Furthermore, using transfer learning based on pre-trained embeddings, seems to result in better performance when compared to end-to-end learning of a typical classifier.

\begin{table*}
\caption{Average precision and F1 scores for the classification tasks using different pre-trained embedding approaches.}
\begin{center}
\begin{tabular}{l|l|l|l|l}
\multirow{2}{*}{ICD-10 code} 																										 & \multicolumn{2}{l|}{Pre-trained embeddings method for:} 			 & \multirow{2}{*}{Average precision} & \multirow{2}{*}{F1-Score} \\
																																								 & Diseases        				& Sex, region and year of birth        &                                    &                           \\\hline \hline 
\multirow{7}{*}{I10: Essential (primary) hypertension} 												 & \multicolumn{2}{l|}{Random initialisation }                   	 & 12.74\%                            & 20.27\%                   \\
																																								 & AE		                	& AE                                   & 8.50\%                             & 12.83\%                  	\\
																																								 & NCF	                	& NCF                                  & 11.86\%                            & 19.30\%                         	\\
																																								 & CBOW	                	& AE	                                 & 14.69\%                            & 22.42\%                         	\\
																																								 & CBOW	                	& NCF	                                 & 15.08\%                  				  & \textbf{22.50\%}                         	\\
																																								 & CBOWA                	& AE	                                 & 13.99\%                            & 21.34\%                         	\\
																																								 & CBOWA                	& NCF	                                 & 14.19\%                            & 21.69\%                         	\\
																																								 & BEHRT                	& AE	                                 & 15.04\%                            & 19.47\%                         	\\
																																								 & BEHRT                	& NCF	                                 & \textbf{15.35\%}                   & 20.29\%                         	\\\hline																																								
																																								
\multirow{7}{*}{M79: Soft Tissue Disorders} 																	 & \multicolumn{2}{l|}{Random initialisation }                   	 & 7.45\%                             & 0.70\%                   \\
																																								 & AE		                	& AE                                   & 5.49\%                             & 9.79\%                  	\\
																																								 & NCF	                	& NCF                                  & 7.32\%                             & 0.77\%                          	\\
																																								 & CBOW	                	& AE	                                 & 8.30\%                             & 13.94\%                         	\\
																																								 & CBOW	                	& NCF	                                 & 8.53\%                   				  & 14.04\%                        	\\
																																								 & CBOWA                	& AE	                                 & 7.62\%                             & 13.16\%                         	\\
																																								 & CBOWA                	& NCF	                                 & 7.97\%                             & 13.54\%                         	\\
																																								 & BEHRT                	& AE	                                 & 8.71\%                             & 14.27\%                         	\\
																																								 & BEHRT                	& NCF	                                 & \textbf{8.74\%}                    & \textbf{14.53\%}                	\\\hline
																																								
\multirow{7}{*}{R10: Abdominal and pelvic pain} 															 & \multicolumn{2}{l|}{Random initialisation }                     & 7.15\%                             & 10.07\%                   \\
																																								 & AE		                	& AE                                   & 5.75\%                             & 10.14\%                  	\\
																																								 & NCF	                	& NCF                                  & 6.81\%                             & 2.65\%                         	\\
																																								 & CBOW	                	& AE	                                 & 7.85\%                             & 13.16\%                         	\\
																																								 & CBOW	                	& NCF	                                 & 8.01\%                    					& 13.22\%                         	\\
																																								 & CBOWA                	& AE	                                 & 7.60\%                             & 12.74\%                         	\\
																																								 & CBOWA                	& NCF	                                 & 7.83\%                             & 13.08\%                         	\\					
																																								 & BEHRT                	& AE	                                 & 8.10\%					                    & 13.57\%                 	\\
																																								 & BEHRT                	& NCF	                                 & \textbf{8.25\%}                    & \textbf{13.77\%}                 	\\
\end{tabular}
\end{center}
\label{tab:prediction_results}
\end{table*}

\subsection{Reliability Analysis}
\begin{figure*}[ht]
\centerline{\includegraphics[width=.7\textwidth]{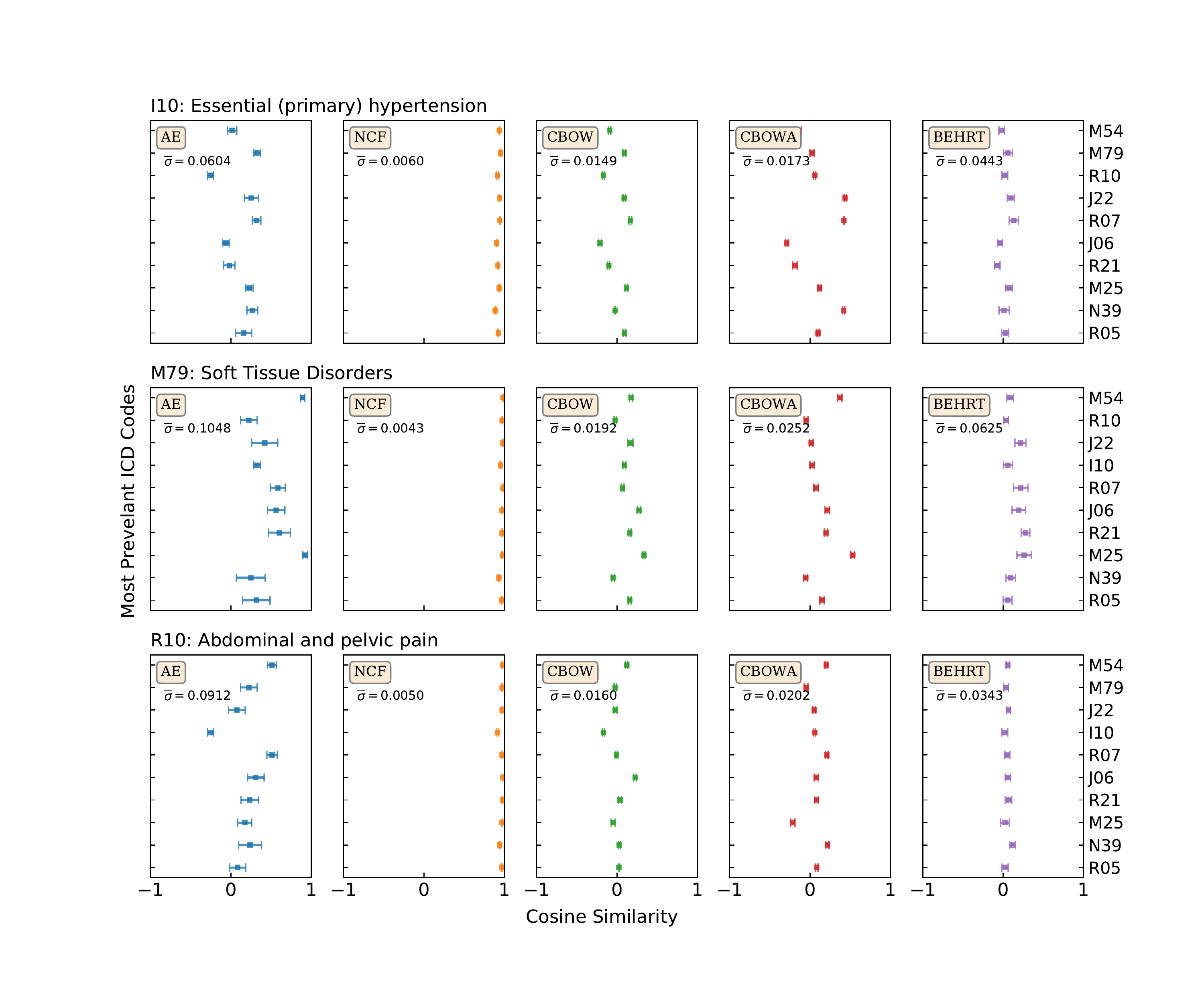}}
\caption{Cosine similarities between the 10 most prevalent ICD-10 codes and I10, M79, R10, respectively. The mean (point) and one standard deviation (error bar) are shown. The average standard deviation for each method ($\sigma$) is also shown.}
\label{fig:reliability1}
\end{figure*}

In order to carry out the first reliability analysis, we selected three of the most common ICD codes (i.e., I10, M79, and R10). For each of these ICD codes, we showed the robustness of their cosine similarity against 10 other codes across 10 different runs (obtained using 10 random splits on the data into training (80\% of patients), validation (20\%) and test (20\%) sets). According to the results (shown in Fig. \ref{fig:reliability1} CBOW, CBOWA and BEHRT result in the most stable embeddings for different levels of similarity. The embeddings resulting from AE seem to have larger variations as one goes from one run to the other; the embeddings from NCF also seems stable, however, it clearly shows that NCF tends to learn highly similar embeddings for the most prevalent ICD codes which is not useful in practice.

\begin{figure}[ht]
\centerline{\includegraphics[width=\columnwidth]{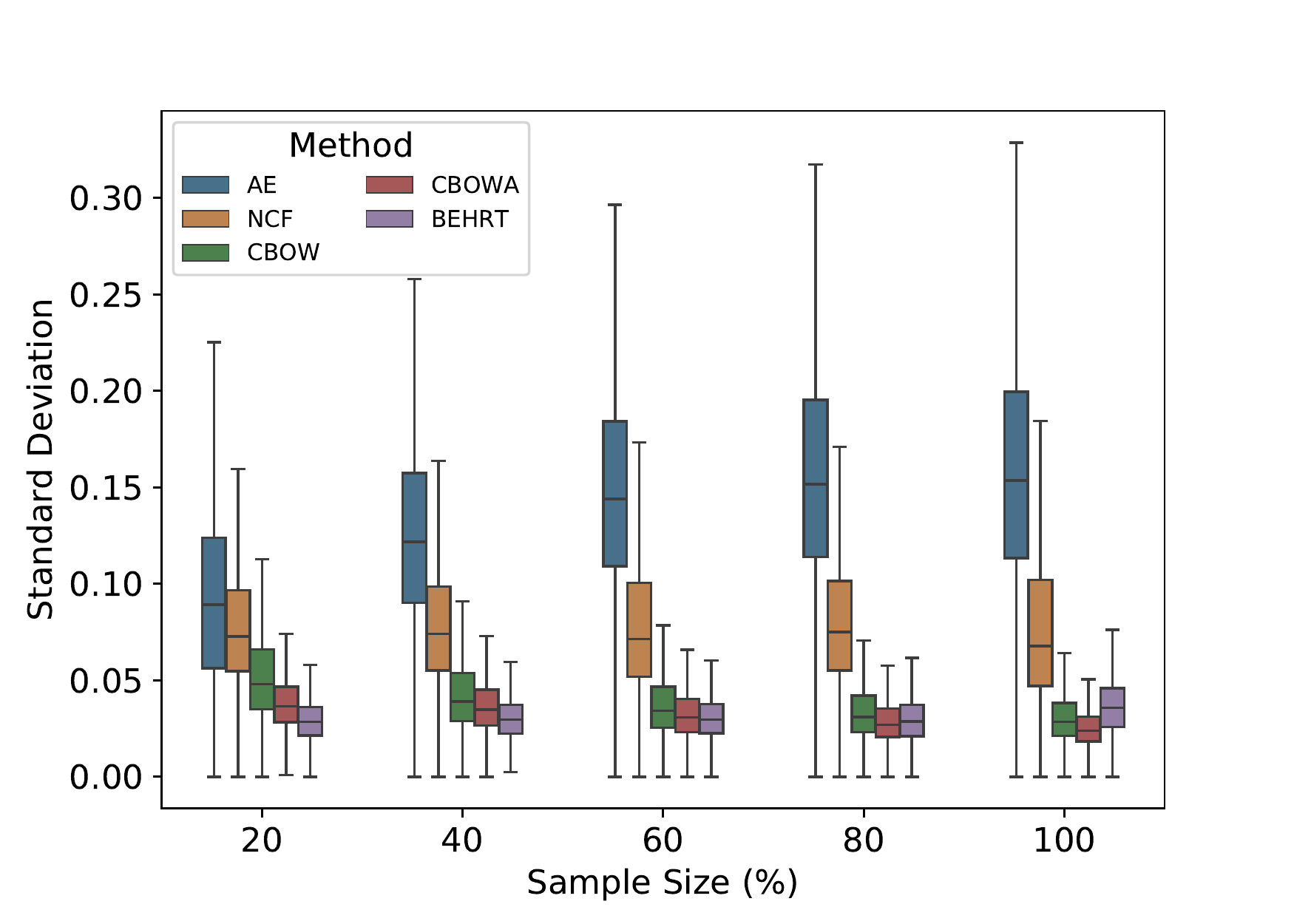}}
\caption{Standard deviation of the embeddings' cosine similarities obtained for varying training sample sizes. Overall, context representation learning methods are more stable.}
\label{fig:reliability2}
\end{figure}

The result of the second reliability analysis (stability of the embeddings as we vary the portion of the data we use for learning them) is shown in Fig. \ref{fig:reliability2}. We trained the models on 20\%, 40\%, 60\%, 80\% and 100\% of the available data (the x-axis) -- 10 times for each case --  and showed the standard deviation of the cosine similarity of the embeddings on the y-axis. In line with all our results so far, contextual representation models (i.e., CBOW, CBOWA and BEHRT) are more reliable.

\section{Discussion}
\label{sec:discussion}
In this study, we investigated the properties of disease embeddings obtained from different neural representation-learning models. There were two primary objectives behind this research: (1) provide a range of frameworks (both qualitative and quantitative) for evaluating such embeddings and comparing them, and (2) encourage the field to share the embeddings for various EHR concepts (ideally, after such evaluations) for various downstream tasks through transfer learning. All the disease embeddings that resulted from our analyses will be shared at https://github.com/deepmedicine/medical-concept-embeddings, which can be used by other researchers to map disease codes to numerical spaces, which are suitable for statistical and algebraic operations.

One of the major contributions of our work was the provision of a set of new quantitative benchmarks for the evaluation of disease embeddings. We used the diagnosis pairs provided by Jensen et al. \cite{jensen2014temporal}, Beam et al. \cite{beam2019clinical}, and Dalianis et al. \cite{dalianis2009stockholm,comorbidity_view}, as clinical insights that our embeddings should show some degree of agreement with. These pairs are extracted via mining large EHR cohorts as well as extensive study of a large body of medical literature. Therefore, the extent of agreement between them and a set of disease embeddings can provide an effective numerical evaluation tool. Our results showed that the context-aware representation-learning models were better at capturing such known relationships, which can explained by how localised information is captured in these models process. Furthermore, their results seem to be more stable going from run to run (i.e., different random initialisations) and changing the training sample sizes (more specifically, changing the proportion of the available data that was used for training); see Fig. \ref{fig:reliability1} and \ref{fig:reliability2}.

Our results, showing that the best embeddings were obtained from the BEHRT architectures, agree with the huge success the Transformer architecture gained in other studies \cite{vaswani2017attention,devlin2018bert,li2020behrt}. So far, the concept of embeddings has been extensively used in NLP, and its applicability has extended to other fields, such as healthcare. As we foresee that this would continue to be the trend -- with new and more advanced neural network architectures being developed -- our approach can be used for assessing such new embedding techniques. Plus, the embeddings that resulted from our analyses are shared with the field so they can be compared against other alternatives and/or used for transfer learning.

\section{Supplementary Information}
\subsection{Data Availability and Reproducibility}
\textbf{CPRD Data Access}: The CPRD database used for this study has been approved by an Independent Scientific Advisory Committee (ISAC). The ISAC protocol number for this study is: $17\_224R$.

To obtain access to CPRD data, researchers are advised to follow the required procedure on the CPRD website Data Access page (https://www.cprd.com/Data-access).

\textbf{Reproducibility}: With regards to the reproducibility of this study, we are committed and pleased to share all the source codes underlying our implementations to empower others to reproduce our work. However, codes related to handling and processing the CPRD data cannot be simply shared as they may reveal sensitive information about the data. More importantly, all research protocols for CPRD data access must be submitted to the ISAC Secretariat using the Protocol Application Form, and if successful, would receive a specific cut of the CPRD data based on the research protocol. We note that, different research protocols would generally receive different data cuts, thus, it is not possible to obtain exactly same benchmark dataset as we used in our study. However, all the major conclusions drew from our work should be reproducible. In order to facilitate such reproducibility, we advocated for minimal data preprocessing and have included all the important data preparation information in our manuscript under the Materials and Methods section.

\subsection*{Cosine similarity metric for 10-nearest neighbours\label{sec:10NN}}
Cosine distance for 10-nearest neighbours are shown in Tables~\ref{tab:Top10-I10}, ~\ref{tab:Top10-M79} and ~\ref{tab:Top10-R10}.

\begin{table*}
\caption{Top 10 diseases closest to I10: Essential (primary) hypertension}
\begin{center}
\begin{tabular}{|c|l|l|}
\hline 
Method & Cosine Similarity & ICD-10 Code\\
\hline 
\hline
\multirow{10}{*}{AE}
& 0.96069 & E78: Disorders of lipoprotein metabolism and other lipidaemias\\
& 0.94603 & Y52: Agents primarily affecting the cardiovascular system\\
& 0.93265 & I11: Hypertensive heart disease\\
& 0.92673 & I70: Atherosclerosis\\
& 0.91515 & I65: Occlusion and stenosis of precerebral arteries, not resulting in cerebral infarction\\
& 0.91113 & R03: Abnormal blood-pressure reading, without diagnosis\\
& 0.90615 & I73: Other peripheral vascular diseases\\
& 0.89762 & I35: Nonrheumatic aortic valve disorders\\
& 0.89018 & I44: Atrioventricular and left bundle-branch block\\
& 0.87448 & I74: Arterial embolism and thrombosis\\
\hline 
\multirow{10}{*}{CBOW}
& 0.50812 & I11: Hypertensive heart disease\\
& 0.47293 & R74: Abnormal serum enzyme levels\\
& 0.44107 & M15: Polyarthrosis\\
& 0.43357 & M10: Gout\\
& 0.41775 & I12: Hypertensive renal disease\\
& 0.41319 & I35: Nonrheumatic aortic valve disorders\\
& 0.40996 & I65: Occlusion and stenosis of precerebral arteries, not resulting in cerebral infarction\\
& 0.4096 & I22: Subsequent myocardial infarction\\
& 0.40455 & H40: Glaucoma\\
& 0.40136 & I44: Atrioventricular and left bundle-branch block\\
\hline 
\multirow{10}{*}{CBOWA}
& 0.78802 & I11: Hypertensive heart disease\\
& 0.77610 & E78: Disorders of lipoprotein metabolism and other lipidaemias\\
& 0.75836 & I20: Angina pectoris\\
& 0.72871 & I25: Chronic ischaemic heart disease\\
& 0.72319 & I12: Hypertensive renal disease\\
& 0.71360 & E11: Type 2 diabetes mellitus\\
& 0.71235 & I65: Occlusion and stenosis of precerebral arteries, not resulting in cerebral infarction\\
& 0.69908 & I22: Subsequent myocardial infarction\\
& 0.69247 & I48: Atrial fibrillation and flutter\\
& 0.69182 & I13: Hypertensive heart and renal disease\\
\hline 
\multirow{10}{*}{NCF}
& 0.95770 & M19: Other arthrosis\\
& 0.95499 & E78: Disorders of lipoprotein metabolism and other lipidaemias\\
& 0.95240 & M79: Other soft tissue disorders, not elsewhere classified\\
& 0.94367 & L98: Other disorders of skin and subcutaneous tissue, not elsewhere classified\\
& 0.94161 & J22: Unspecified acute lower respiratory infection\\
& 0.94026 & M54: Dorsalgia\\
& 0.93908 & R03: Abnormal blood-pressure reading, without diagnosis\\
& 0.93847 & R07: Pain in throat and chest\\
& 0.93586 & R06: Abnormalities of breathing\\
& 0.93153 & K62: Other diseases of anus and rectum\\
\hline 
\multirow{10}{*}{BEHRT}
& 0.27687 & X61: Intentional self-poisoning by and exposure to antiepileptic, sedative-hypnotic...\\
& 0.25725 & Q90: Down syndrome\\
& 0.25159 & H40: Glaucoma\\
& 0.24646 & G40: Epilepsy\\
& 0.24359 & J44: Other chronic obstructive pulmonary disease\\
& 0.23918 & E03: Other hypothyroidism\\
& 0.22216 & F32: Depressive episode\\
& 0.22018 & M13: Other arthritis\\
& 0.21731 & F20: Schizophrenia\\
& 0.21652 & E78: Disorders of lipoprotein metabolism and other lipidaemias\\
\hline 
\end{tabular}
\end{center}
\label{tab:Top10-I10}
\end{table*}

\begin{table*}
\caption{Top 10 diseases closest to M79: Other soft tissue disorders, not elsewhere classified}
\begin{center}
\begin{tabular}{|c|l|l|}
\hline 
Method & Cosine Similarity & ICD-10 Code\\
\hline 
\hline
\multirow{10}{*}{AE}
& 0.90472 & M71: Other bursopathies\\
& 0.90431 & G57: Mononeuropathies of lower limb\\
& 0.89115 & M65: Synovitis and tenosynovitis\\
& 0.88924 & R20: Disturbances of skin sensation\\
& 0.88842 & S33: Dislocation, sprain and strain of joints and ligaments of lumbar spine and pelvis\\
& 0.87179 & M02: Reactive arthropathies\\
& 0.86672 & M25: Other joint disorders, not elsewhere classified\\
& 0.8646 & M75: Shoulder lesions\\
& 0.86387 & M76: Enthesopathies of lower limb, excluding foot\\
& 0.86325 & M70: Soft tissue disorders related to use, overuse and pressure\\
\hline 
\multirow{10}{*}{CBOW}
& 0.47778 & M70: Soft tissue disorders related to use, overuse and pressure\\
& 0.46388 & M65: Synovitis and tenosynovitis\\
& 0.44834 & M77: Other enthesopathies\\
& 0.44325 & M67: Other disorders of synovium and tendon\\
& 0.42256 & L84: Corns and callosities\\
& 0.42254 & M76: Enthesopathies of lower limb, excluding foot\\
& 0.4146 & M71: Other bursopathies\\
& 0.41433 & G57: Mononeuropathies of lower limb\\
& 0.39715 & R52: Pain, not elsewhere classified\\
& 0.38953 & S46: Injury of muscle and tendon at shoulder and upper arm level\\
\hline 
\multirow{10}{*}{CBOWA}
& 0.59134 & I83: Varicose veins of lower extremities\\
& 0.55397 & M71: Other bursopathies\\
& 0.54512 & G57: Mononeuropathies of lower limb\\
& 0.54505 & M76: Enthesopathies of lower limb, excluding foot\\
& 0.53288 & M65: Synovitis and tenosynovitis\\
& 0.52307 & M20: Acquired deformities of fingers and toes\\
& 0.51443 & M66: Spontaneous rupture of synovium and tendon\\
& 0.50651 & M70: Soft tissue disorders related to use, overuse and pressure\\
& 0.50574 & M25: Other joint disorders, not elsewhere classified\\
& 0.49245 & L84: Corns and callosities\\
\hline 
\multirow{10}{*}{NCF}
& 0.98563 & M54: Dorsalgia\\
& 0.9786 & M25: Other joint disorders, not elsewhere classified\\
& 0.97852 & R10: Abdominal and pelvic pain\\
& 0.97815 & R07: Pain in throat and chest\\
& 0.97807 & J22: Unspecified acute lower respiratory infection\\
& 0.97252 & R21: Rash and other nonspecific skin eruption\\
& 0.9709 & J06: Acute upper respiratory infections of multiple and unspecified sites\\
& 0.97035 & R05: Cough\\
& 0.96738 & K30: Functional dyspepsia\\
& 0.96652 & L08: Other local infections of skin and subcutaneous tissue\\
\hline 
\multirow{10}{*}{BEHRT}
& 0.30496 & R21: Rash and other nonspecific skin eruption\\
& 0.30296 & R07: Pain in throat and chest\\
& 0.269 & L98: Other disorders of skin and subcutaneous tissue, not elsewhere classified\\
& 0.2646 & K30: Functional dyspepsia\\
& 0.26124 & M25: Other joint disorders, not elsewhere classified\\
& 0.24571 & L08: Other local infections of skin and subcutaneous tissue\\
& 0.21928 & R22: Localized swelling, mass and lump of skin and subcutaneous tissue\\
& 0.21867 & B35: Dermatophytosis\\
& 0.20065 & T14: Injury of unspecified body region\\
& 0.19411 & J06: Acute upper respiratory infections of multiple and unspecified sites\\
\hline 
\end{tabular}
\end{center}
\label{tab:Top10-M79}
\end{table*}

\begin{table*}
\caption{Top 10 diseases closest to R10: Abdominal and pelvic pain}
\begin{center}
\begin{tabular}{|c|l|l|}
\hline 
Method & Cosine Similarity & ICD-10 Code\\
\hline 
\hline
\multirow{10}{*}{AE}
& 0.91692 & K58: Irritable bowel syndrome\\
& 0.90787 & R12: Heartburn\\
& 0.8933 & K66: Other disorders of peritoneum\\
& 0.88993 & B80: Enterobiasis\\
& 0.88897 & K81: Cholecystitis\\
& 0.88667 & I84: Haemorrhoids\\
& 0.88505 & K50: Crohn disease [regional enteritis]\\
& 0.88488 & K82: Other diseases of gallbladder\\
& 0.88467 & B98: Other specified infectious agents as the cause of diseases classified to other chapters\\
& 0.88383 & R76: Other abnormal immunological findings in serum\\
\hline 
\multirow{10}{*}{CBOW}
& 0.38417 & R12: Heartburn\\
& 0.37309 & R14: Flatulence and related conditions\\
& 0.34671 & K82: Other diseases of gallbladder\\
& 0.31971 & K66: Other disorders of peritoneum\\
& 0.31593 & R19: Other symptoms and signs involving the digestive system and abdomen\\
& 0.30709 & G43: Migraine\\
& 0.30615 & E73: Lactose intolerance\\
& 0.30376 & O21: Excessive vomiting in pregnancy\\
& 0.29091 & J02: Acute pharyngitis\\
& 0.28959 & N97: Female infertility\\
\hline 
\multirow{10}{*}{CBOWA}
& 0.67008 & R19: Other symptoms and signs involving the digestive system and abdomen\\
& 0.65309 & R14: Flatulence and related conditions\\
& 0.63626 & K83: Other diseases of biliary tract\\
& 0.60278 & K37: Unspecified appendicitis\\
& 0.60188 & K66: Other disorders of peritoneum\\
& 0.60182 & Q43: Other congenital malformations of intestine\\
& 0.60083 & K82: Other diseases of gallbladder\\
& 0.5987 & K38: Other diseases of appendix\\
& 0.59438 & K81: Cholecystitis\\
& 0.58445 & Q44: Congenital malformations of gallbladder, bile ducts and liver\\
\hline 
\multirow{10}{*}{NCF}
& 0.98303 & M54: Dorsalgia\\
& 0.98138 & R21: Rash and other nonspecific skin eruption\\
& 0.97895 & J06: Acute upper respiratory infections of multiple and unspecified sites\\
& 0.97852 & M79: Other soft tissue disorders, not elsewhere classified\\
& 0.97846 & J22: Unspecified acute lower respiratory infection\\
& 0.97568 & M25: Other joint disorders, not elsewhere classified\\
& 0.97487 & R05: Cough\\
& 0.97297 & R07: Pain in throat and chest\\
& 0.9711 & L08: Other local infections of skin and subcutaneous tissue\\
& 0.96679 & H60: Otitis externa\\
\hline 
\multirow{10}{*}{BEHRT}
& 0.24785 & D64: Other anaemias\\
& 0.18252 & K62: Other diseases of anus and rectum\\
& 0.18163 & D50: Iron deficiency anaemia\\
& 0.17432 & K52: Other noninfective gastroenteritis and colitis\\
& 0.1595 & R33: Retention of urine\\
& 0.15783 & D69: Purpura and other haemorrhagic conditions\\
& 0.15201 & K92: Other diseases of digestive system\\
& 0.15122 & K83: Other diseases of biliary tract\\
& 0.15053 & R12: Heartburn\\
& 0.14959 & K21: Gastro-oesophageal reflux disease\\
\hline  
\end{tabular}
\end{center}
\label{tab:Top10-R10}
\end{table*}



\section*{Acknowledgment}
 \clearpage
This work used data provided by patients and collected by the NHS as part of their care and support and would not have been possible without access to this data. The NIHR recognises and values the role of patient data, securely accessed and stored, both in underpinning and leading to improvements in research and care.
We also thank Wayne Dorrington for his help in making Fig. \ref{fig:architectures}.

\bibliographystyle{IEEEtran} 
\bibliography{bib}

\end{document}